%% file: main.tex
\documentclass[10pt,twocolumn,letterpaper]{article}
\usepackage[pagenumbers]{cvpr}
\input{preamble}
\definecolor{cvprblue}{rgb}{0.21,0.49,0.74}
\usepackage[pagebackref,breaklinks,colorlinks,allcolors=cvprblue]{hyperref}
\def\confName{CVPR}
\def\confYear{2027}

\title{Rethinking Unsupervised Cross-Modal Homography Estimation:\\
Spatially Aware Representations and Geometric Consistency Learning}
\author{%
{\large\normalfont Zhuoran Yang$^{1}$\quad Junchen Yu$^{1}$\quad Runmin Zhang$^{1}$\quad Linshan Li$^{1}$}\\[3pt]
{\large\normalfont Chenghao Zhang$^{1}$\quad Hui-liang Shen$^{1,4}$\quad Si-Yuan Cao$^{2,3,4}$}\\[5pt]
{\large\normalfont $^{1}$College of Information Science and Electronic Engineering, Zhejiang University}\\[2pt]
{\large\normalfont $^{2}$Ningbo Global Innovation Center, Zhejiang University\qquad $^{3}$NingboTech University}\\[2pt]
{\large\normalfont $^{4}$Jinhua Institute, Zhejiang University}\\[3pt]
{\ttfamily\small yangzhr2025@zju.edu.cn}
}

\begin{document}
\maketitle
\suppressfloats[t]
\input{sec/0_abstract}
\input{sec/1_intro}
\input{sec/2_related_work}
\input{sec/3_method}
\input{sec/4_experiments}
\input{sec/5_conclusion}

{\small
\bibliographystyle{ieeenat_fullname}
\bibliography{references}
}
\end{document}

%% file: preamble.tex
\newcommand{\Ltrans}{\ensuremath{\mathcal{L}_{\mathrm{trans}}}}
\newcommand{\Lhomo}{\ensuremath{\mathcal{L}_{\mathrm{homo}}}}
\DeclareMathOperator{\sg}{sg}

%% file: sec/0_abstract.tex
\begin{abstract}
Unsupervised cross-modal homography estimation is challenging because appearance differences obscure shared geometric structures. Separating modality transfer from homography estimation helps reduce these differences, but accurate alignment still requires effective geometric representations and consistency constraints on the two subtasks. We propose GeoCo (Geometry-Aware and Consistency-Constrained Homography Estimation), a framework that addresses these two requirements without cross-modal transformation labels. Its homography estimator, RVHomo, adapts hierarchical features from a frozen vision foundation model (VFM) and fuses them with local CNN features to capture structural context and spatial detail. Two geometric consistency constraints encourage modality transfer to preserve geometry and homography predictions to remain consistent under synthetic geometric transformations. Through two-stage progressive distillation, we further obtain GeoCo-D, which uses the lightweight RVHomo-D estimator. An intermediate model first learns to perform cross-modal matching without modality transfer, then transfers prediction and feature knowledge to the student, eliminating online VFM extraction. Experiments on GoogleMap, OPT-SAR, FSD, and VIS-IR show that GeoCo reduces mean average corner error by 17.4--46.7\% relative to the best-performing baseline on each dataset. GeoCo-D also outperforms all compared baselines while using 98.5\% fewer parameters and 70.4\% fewer FLOPs than the complete GeoCo teacher.
\end{abstract}

%% file: sec/1_intro.tex
\section{Introduction}
\label{sec:intro}

\begin{figure}[t]
    \centering
    \includegraphics[width=\linewidth]{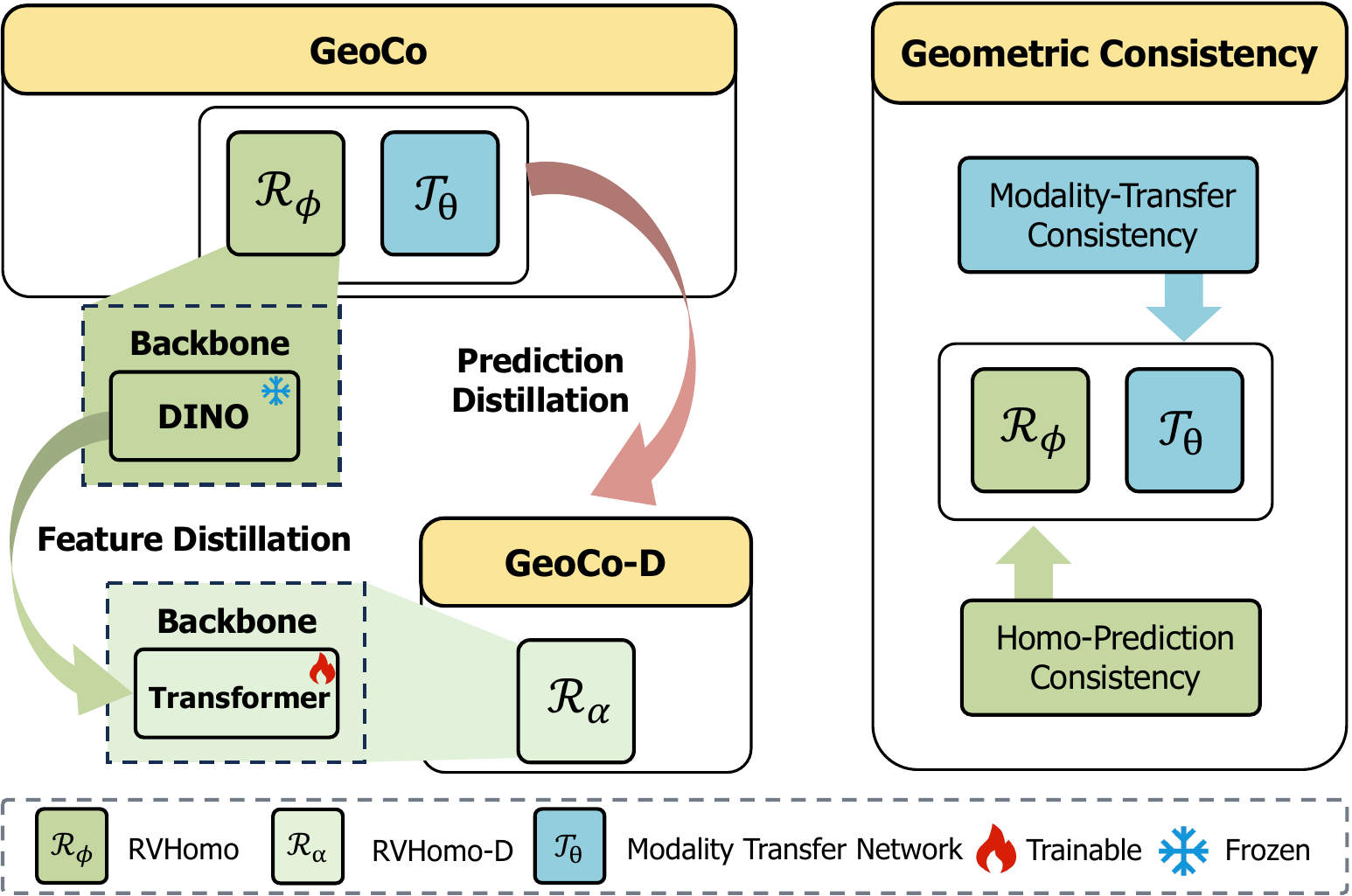}
    \caption{GeoCo enhances the SSHNet unsupervised framework with the RVHomo estimator and geometric consistency constraints. Progressive distillation produces RVHomo-D without modality transfer or online VFM extraction.}
    \label{fig:teaser}
\end{figure}

Homography estimation recovers the global projective transformation between two views for image registration, multi-sensor fusion, and localization. Across imaging modalities, the same scene structures can have substantially different appearances, making spatial correspondences difficult to establish~\cite{heinrich2012mind,kim2015dasc,li2020rift,pielawski2020comir}. Without cross-modal transformation labels for training, a model must learn to distinguish geometric misalignment from appearance differences without direct supervision of the desired transformation.

Existing unsupervised methods construct learning signals from image reconstruction, feature similarity, or synthetic transformations within each modality~\cite{nguyen2018unsupervised,zhang2020content,koguciuk2021perceptual,zhang2024scpnet}. However, cross-modal appearance differences complicate supervision based on image or feature discrepancies. To address this challenge, cross-modal methods have explored alternating optimization of representations and geometry~\cite{song2024alto}, as well as learning strategies that incorporate modality transfer~\cite{arar2020nemar,yu2025sshnet}. Among them, SSHNet~\cite{yu2025sshnet} reformulates the problem into modality transfer and homography estimation and trains the two components through split optimization. Modality transfer reduces appearance discrepancies, while intra-modal geometric supervision supports learning the estimator.

We revisit geometric learning within SSHNet's decomposed framework, focusing on two aspects: informative and precisely localized structural representations, and explicit constraints on how each subtask responds to geometric transformations.

First, SSHNet reduces appearance differences through modality transfer, but its homography estimator still needs effective geometric representations for accurate alignment. To strengthen geometric representations within this framework, we introduce a pretrained vision foundation model (VFM)~\cite{oquab2023dinov2}, whose broad structural context complements the local details captured by CNN features. We adapt and integrate these complementary features using SSHNet's self-supervised learning signals, enabling more precise homography estimation without cross-modal transformation labels.

Second, the existing objectives do not explicitly enforce how the two subtasks should respond to known geometric transformations. Modality transfer changes appearance, so transforming an input before transfer should produce a result consistent with transforming the transferred image afterward. Homography estimation predicts spatial relationships, so a known transformation of the source image should induce a corresponding change in the prediction according to homography composition. These relations provide additional geometric constraints without requiring transformation labels between the original cross-modal images.

Based on these observations, we propose GeoCo (Geometry-Aware and Consistency-Constrained Homography Estimation), a framework designed to make better use of geometric information for unsupervised cross-modal homography estimation. As illustrated in \cref{fig:teaser}, GeoCo enhances geometric representations through its RVHomo (Recurrent VFM-guided Homography Estimation Network) estimator and introduces geometric consistency constraints for modality transfer and homography prediction. These designs improve how the framework represents geometric structures and learns from their transformation relationships.

The enhanced framework retains modality transfer and introduces the cost of VFM feature extraction. Through two-stage progressive distillation, we obtain GeoCo-D, a lightweight framework that uses RVHomo-D as its homography estimator. As shown in \cref{fig:teaser}, following the prediction-distillation strategy of SSHNet-D~\cite{yu2025sshnet}, the complete GeoCo teacher first supervises an intermediate model that directly processes the original cross-modal pair, removing the need for modality transfer. The intermediate model then transfers prediction and branch-wise feature knowledge to RVHomo-D, whose lightweight Transformer replaces the VFM. This intermediate stage enables feature distillation between models performing the same cross-modal matching task. The resulting GeoCo-D requires neither modality transfer nor online VFM extraction at inference.

Experiments on four cross-modal benchmarks demonstrate the effectiveness of the proposed improvements. GeoCo reduces mean average corner error (MACE) by 17.4--46.7\% relative to the strongest evaluated competitor on each dataset. GeoCo-D also outperforms these competitors while using 98.5\% fewer total parameters and 70.4\% less computation than the complete teacher. Our main contributions are:
\begin{itemize}
    \item We propose GeoCo to enhance geometric learning within SSHNet's decomposed unsupervised framework. Its RVHomo estimator adapts hierarchical VFM features into scale-aligned geometric representations and fuses them with local CNN features without cross-modal transformation labels.
    \item We introduce modality-transfer and homography-prediction consistency constraints that regulate each subtask's response to known geometric transformations while preserving split optimization.
    \item We develop progressive distillation that transfers predictions and representations through an intermediate model to a lightweight student without modality transfer or online VFM feature extraction.
\end{itemize}

%% file: sec/2_related_work.tex
\section{Related Work}

\paragraph{Unsupervised cross-modal homography estimation.}
Unsupervised homography estimators derive supervision from photometric reconstruction, learned features, perceptual similarity~\cite{johnson2016perceptual}, or motion representations~\cite{nguyen2018unsupervised,zhang2020content,koguciuk2021perceptual,ye2021motion,hong2022homogan,liu2023geoformer}. Cross-modal appearance differences confound these signals with geometric misalignment. SCPNet uses intra-modal self-supervision and consistent feature projection~\cite{zhang2024scpnet}, AltO alternates representation and geometry optimization~\cite{song2024alto}, and SSHNet separates modality transfer from homography estimation through split optimization~\cite{yu2025sshnet}. SSHNet reduces appearance gaps through transfer and supervises its estimator with synthetic intra-modal transformations. This decomposition provides a learning framework, but precise alignment still depends on spatially informative features and geometrically consistent subtask responses. GeoCo retains SSHNet's decomposition, objectives, and optimization while improving representation adaptation and imposing subtask-specific consistency constraints without cross-modal transformation labels.

\paragraph{VFM-based matching and homography estimation.}
Self-supervised pretraining provides structural representations~\cite{caron2021dino,weinzaepfel2022croco}, with pretrained features used for multimodal registration and semantic correspondence~\cite{song2024dinoreg,tang2023dift,zhang2023twofeatures}. To support precise localization, RoMa combines deepest-layer DINOv2 features for coarse matching with CNN refinement~\cite{edstedt2024roma}; GFNet applies cross-view DINOv2 interaction and conditions fine CNN features on coarse representations~\cite{zhang2025gfnet}; DistillMatch distills VFM knowledge into a lightweight multimodal matcher~\cite{yang2025distillmatch}. These methods exploit complementary features, with adaptation tailored to matching pipelines trained under explicit geometric supervision. Style-diverse synthesis and color-invariant representations address unseen modality combinations~\cite{you2026generalized}. Within SSHNet, the challenge is to adapt pretrained features for precise multi-scale estimation using the available unsupervised learning signals. To address this challenge, we design RVHomo as the homography estimator within GeoCo, enabling the framework to make better use of geometric information in VFM features.

%% file: sec/3_method.tex
\section{Method}
\label{sec:method}

\subsection{Overview}

\begin{figure}[!ht]
    \centering
    \includegraphics[width=\linewidth]{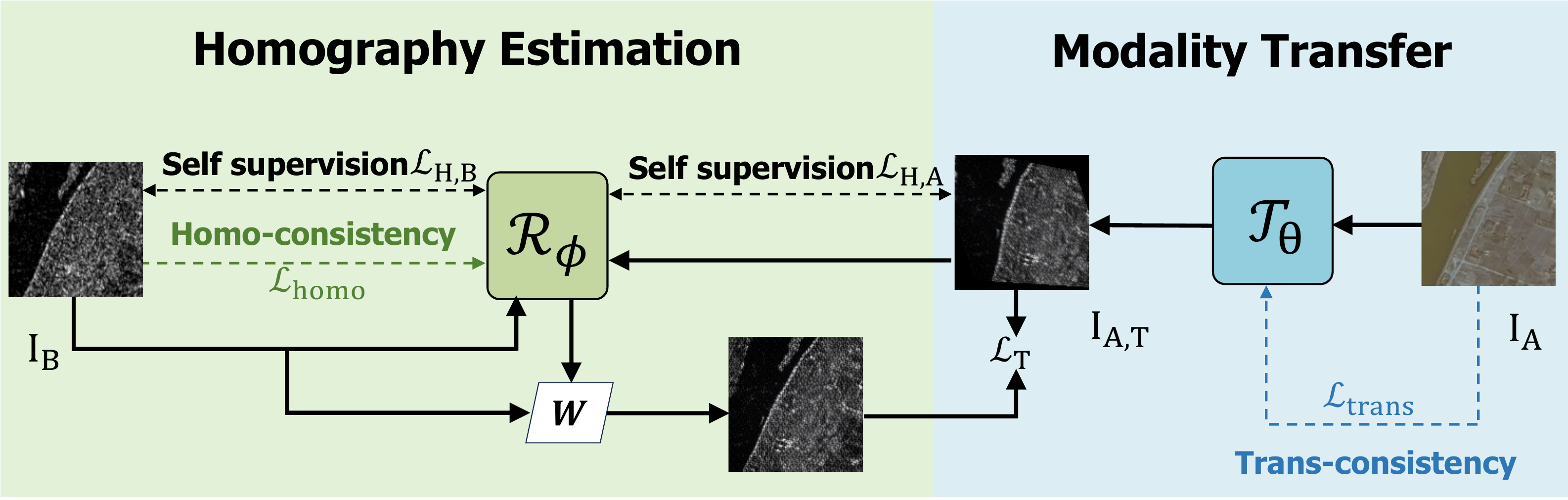}
    \caption{Overview of GeoCo. The framework retains SSHNet's decomposition into homography estimation and modality transfer with split optimization. RVHomo ($\mathcal R_\phi$) enhances geometric representations, while $\mathcal L_{\mathrm{homo}}$ and $\mathcal L_{\mathrm{trans}}$ constrain the geometric behavior of the estimator and modality-transfer network ($\mathcal T_\theta$), respectively.}
    \label{fig:overview}
\end{figure}

GeoCo retains SSHNet's problem reformulation, original learning objectives, and split optimization~\cite{yu2025sshnet}, while enhancing geometric representations and imposing consistency constraints, as shown in \cref{fig:overview}. RVHomo serves as the homography estimator in GeoCo. Given source and target images $\mathbf I_{\mathrm A}$ and $\mathbf I_{\mathrm B}$ from different modalities, it fuses local CNN and adapted VFM features for coarse-to-fine recurrent prediction of four-corner displacements. Two geometric consistency constraints regularize modality transfer and homography prediction using self-supervision from synthetic geometric perturbations, without requiring cross-modal transformation labels.

\subsection{RVHomo: Geometry-Aware Representation Learning}

\begin{figure*}[t]
    \centering
    \includegraphics[width=\textwidth]{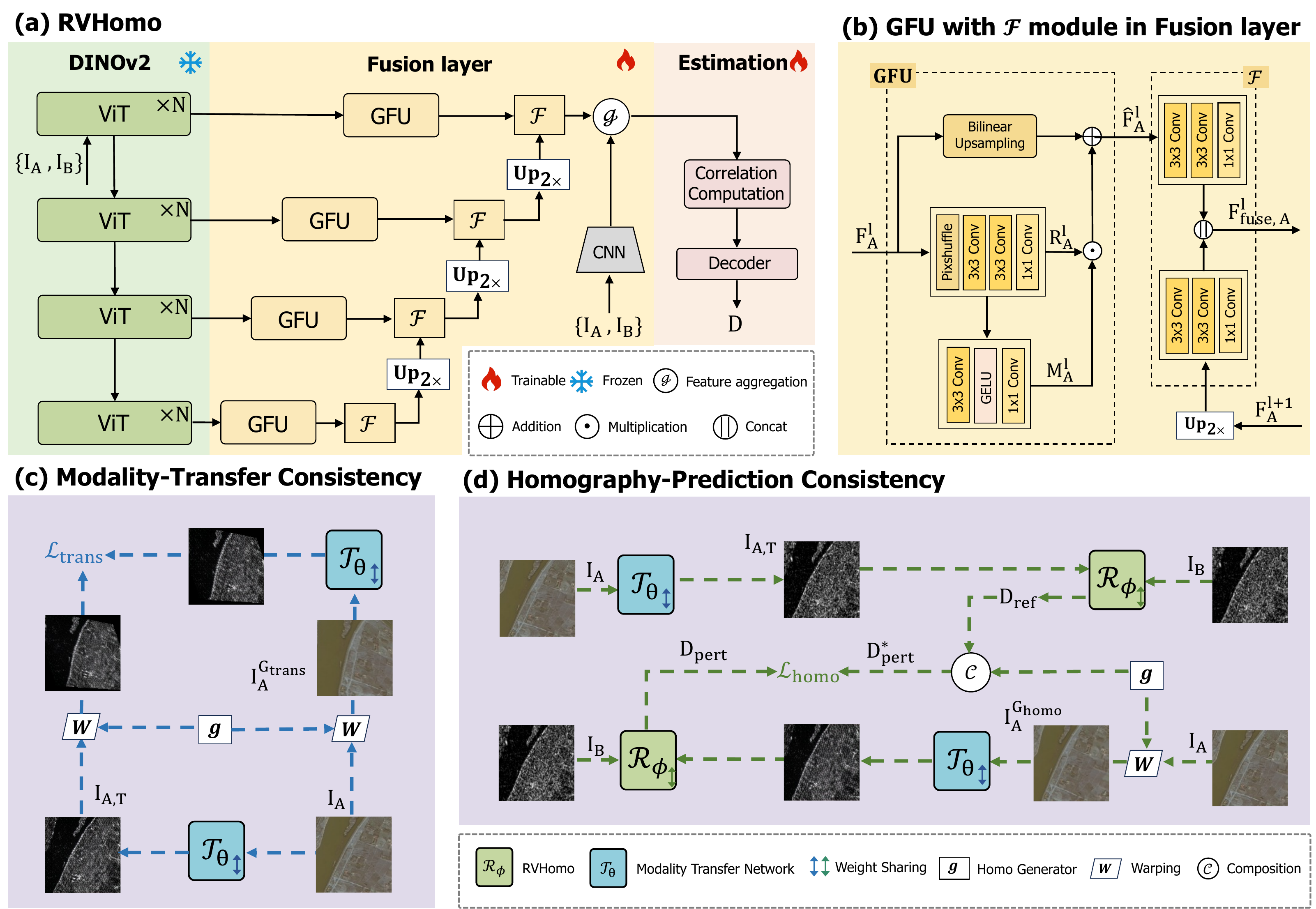}
    \caption{RVHomo and geometric consistency constraints. (a) RVHomo fuses frozen DINOv2 and local CNN features for multi-scale homography estimation. (b) GFU and top-down fusion adapt hierarchical VFM features into scale-aligned geometric representations. (c) Modality-transfer consistency constrains geometry preservation during appearance translation. (d) Homography-prediction consistency constrains predictions under known source-side perturbations.}
    \label{fig:method-details}
\end{figure*}

As detailed in \Cref{fig:method-details}(a,b), a CNN--VFM dual-branch encoder is shared between the source and target images and produces matching features at scales $s\in\{\tfrac14,\tfrac12\}$. For clarity, we describe the source branch for $\mathbf I_{\mathrm{A}}$; the target image $\mathbf I_{\mathrm{B}}$ follows the same operations with shared weights. The CNN branch extracts local image structures, while the frozen VFM partitions $\mathbf I_{\mathrm{A}}$ into non-overlapping patches and processes the resulting token sequence with a Transformer encoder. After removing non-spatial tokens, the outputs of $L$ selected Transformer blocks, indexed by $\{\ell_j\}_{j=1}^{L}$, are linearly projected and spatially rearranged into feature maps $\{\mathbf F_{\mathrm{A}}^{\ell_j}\}_{j=1}^{L}$.

These low-resolution features are transformed to the resolutions used for homography matching by geometry-aware feature upsampling (GFU). Given $\mathbf F_{\mathrm{A}}^{\ell_j}$ from block $\ell_j$ and its upsampling factor $r_j$, GFU adds a gated geometry-refinement residual to a bilinearly upsampled feature:
\begin{equation}
\begin{aligned}
\widehat{\mathbf F}_{\mathrm{A}}^{\ell_j}
&=\operatorname{GFU}(\mathbf F_{\mathrm{A}}^{\ell_j},r_j)\\
&=
\operatorname{BilUp}_{r_j}(\mathbf F_{\mathrm{A}}^{\ell_j})
+
\mathbf M_{\mathrm{A}}^{\ell_j}\odot\mathbf R_{\mathrm{A}}^{\ell_j}.
\end{aligned}
\label{eq:gfu}
\end{equation}
The spatial gate $\mathbf M_{\mathrm{A}}^{\ell_j}$ controls where and how strongly the geometry-refinement residual $\mathbf R_{\mathrm{A}}^{\ell_j}$ is injected. The transformed features are then aggregated along a top-down pathway:
\begin{equation}
\mathbf F_{\mathrm{fuse},\mathrm{A}}^{\ell_j}
=
\mathcal F_j\!\left(
\operatorname{Up}_{2}
\big(\mathbf F_{\mathrm{fuse},\mathrm{A}}^{\ell_{j+1}}\big),
\widehat{\mathbf F}_{\mathrm{A}}^{\ell_j}
\right).
\label{eq:vfm-aggregation}
\end{equation}
For $j=L-1,\ldots,1$, $\mathcal F_j$ transforms the two inputs separately, concatenates them along the channel dimension, and applies further refinement. Scale-specific channel projections yield the aggregated features $\mathbf F_{\mathrm{VFM},\mathrm{A}}^{s}$ at $1/4$ and $1/2$ resolution. Each feature is concatenated with its CNN counterpart and passed through a scale-specific fusion unit:
\begin{equation}
\mathbf F_{\mathrm{Homo},\mathrm{A}}^{s}
=
\mathcal G^{s}\!\left(
\mathbf F_{\mathrm{CNN},\mathrm{A}}^{s}
\mathbin\Vert
\mathbf F_{\mathrm{VFM},\mathrm{A}}^{s}
\right),
\label{eq:fusion}
\end{equation}
where $s\in\{\tfrac14,\tfrac12\}$ and $\Vert$ denotes channel-wise concatenation. Applying the same pipeline to $\mathbf I_{\mathrm{B}}$ yields $\mathbf F_{\mathrm{Homo},\mathrm{B}}^{s}$. The fused source and target features are then processed by the recurrent homography estimation pipeline of IHN~\cite{cao2022iterative}, producing four-corner displacements in a coarse-to-fine manner. We leave its recurrent refinement and prediction head unchanged.

During training, we use auxiliary reconstruction at both matching scales to regularize the aggregated VFM features. Let $\Psi_{4\times}$ and $\Psi_{2\times}$ denote lightweight decoders built with sub-pixel upsampling~\cite{shi2016realtime}. The reconstruction loss is
\begin{equation}
\begin{aligned}
\mathcal L_{\mathrm{recon},\mathrm{A}}
&=
\left\|
\mathbf I_{\mathrm{A}}-\Psi_{4\times}
(\mathbf F_{\mathrm{VFM},\mathrm{A}}^{1/4})
\right\|_2^2\\
&\quad+
\left\|
\mathbf I_{\mathrm{A}}-\Psi_{2\times}
(\mathbf F_{\mathrm{VFM},\mathrm{A}}^{1/2})
\right\|_2^2,\\
\mathcal L_{\mathrm{recon}}
&=
\mathcal L_{\mathrm{recon},\mathrm{A}}
+
\mathcal L_{\mathrm{recon},\mathrm{B}},
\end{aligned}
\label{eq:lrecon}
\end{equation}
where $\mathcal L_{\mathrm{recon},\mathrm{B}}$ is defined identically for $\mathbf I_{\mathrm{B}}$. 

We use blocks $[6,12,18,24]$ of DINOv2-L/14, with one-based indices. The entire VFM remains frozen; the adapters, aggregation, fusion, and estimator learn through SSHNet's objectives and the added regularizers. Reconstruction uses each input as its own target and does not supply cross-modal correspondence labels. Its decoders are discarded at inference.

\subsection{Geometric Consistency Constraints}
\label{sec:lconsist}

\Cref{fig:method-details}(c,d) illustrates the two consistency constraints.
The split optimization of SSHNet decouples modality transfer from homography estimation, but does not explicitly constrain how either process should respond to a known geometric perturbation. Related warp-consistency objectives use known transformations to supervise correspondence predictions without real-image ground truth~\cite{truong2021warp,zhang2026dcflow}. We independently sample two known source-side perturbations $\mathbf G_{\mathrm{trans}},\mathbf G_{\mathrm{homo}}\in\mathbb R^{4\times2}$ for modality-transfer consistency and homography-prediction consistency, respectively. Each perturbation is represented by displacements of the four source corners. For either perturbation $\mathbf G$, let $\mathbf I_{\mathrm{A}}^{\mathbf G}$ denote the resulting source view and let $\mathcal W(\mathbf X;\mathbf G)$ apply the corresponding homography warp to $\mathbf X$. Each $\mathbf G$ is oriented from its perturbed source view to the reference source view.

\paragraph{Modality-Transfer Consistency.}
Let $\mathcal T_\theta$ denote the modality transfer network, and write $\widetilde{\mathbf I}_{\mathrm A}=\mathcal T_\theta(\mathbf I_{\mathrm A})$ and $\widetilde{\mathbf I}_{\mathrm A}^{\mathbf G}=\mathcal T_\theta(\mathbf I_{\mathrm A}^{\mathbf G})$. We compare transfer after warping with warping after transfer. Using the binary validity mask $\mathbf V_{\mathrm{trans}}$ induced by $\mathbf G_{\mathrm{trans}}$, we define
\begin{equation}
\begin{aligned}
\mathcal L_{\mathrm{trans}}
={}&
\frac{1}{\left\|\mathbf V_{\mathrm{trans}}\right\|_1}
\Big\|\mathbf V_{\mathrm{trans}}\odot\Big[
\widetilde{\mathbf I}_{\mathrm A}^{\mathbf G_{\mathrm{trans}}}\\
&\quad-
\sg\!\Big(
\mathcal W(\widetilde{\mathbf I}_{\mathrm A};\mathbf G_{\mathrm{trans}})
\Big)
\Big]\Big\|_1 ,
\end{aligned}
\label{eq:ltrans}
\end{equation}
where $\sg(\cdot)$ denotes stop-gradient. The warped reference transfer serves only as the target; consequently, $\mathcal L_{\mathrm{trans}}$ updates $\mathcal T_\theta$ only through the perturbed-view branch.

\paragraph{Homography-Prediction Consistency.}
Let $\mathcal R_\phi$ denote RVHomo, whose final output is a four-corner displacement. The reference prediction, perturbed-view prediction, and composed pseudo target are
\begin{equation}
\begin{aligned}
\mathbf D_{\mathrm{ref}}
&=
\sg\!\left[
\mathcal R_\phi\!\left(\widetilde{\mathbf I}_{\mathrm A},\mathbf I_{\mathrm{B}}\right)
\right],\\
\mathbf D_{\mathrm{pert}}
&=
\mathcal R_\phi\!\left(
\sg\!\left[\widetilde{\mathbf I}_{\mathrm A}^{\mathbf G_{\mathrm{homo}}}\right],\mathbf I_{\mathrm{B}}
\right),\\
\mathbf D_{\mathrm{pert}}^{\star}
&=
\mathcal C\!\left(\mathbf D_{\mathrm{ref}},\mathbf G_{\mathrm{homo}}\right),
\end{aligned}
\label{eq:prediction-composition}
\end{equation}
where all displacements lie in $\mathbb R^{4\times2}$. Let $\Pi$ convert corner displacements into a homography matrix and $\Pi^{-1}$ convert back. The composition is
\begin{equation}
\mathcal C(\mathbf D,\mathbf G)
=\Pi^{-1}\!\left[\Pi(\mathbf D)\Pi(\mathbf G)\right].
\label{eq:composition}
\end{equation}
Here, $\Pi(\mathbf G)$ maps the perturbed source to the reference source, and $\Pi(\mathbf D)$ maps the reference source to the target. Their product therefore maps the perturbed source directly to the target; corner displacements must not be added element-wise. We then define
\begin{equation}
\mathcal L_{\mathrm{homo}}
=
\operatorname{mean}\!\left(
\left\lvert
\mathbf D_{\mathrm{pert}}
-
\mathbf D_{\mathrm{pert}}^{\star}
\right\rvert
\right),
\label{eq:lhomo}
\end{equation}
where the mean is taken over valid samples and the eight coordinates of the four-corner displacement. The reference prediction $\mathbf D_{\mathrm{ref}}$ and the modality-transferred perturbed input are stop-gradient, so the composed target $\mathbf D_{\mathrm{pert}}^\star$ is also detached. Hence, $\mathcal L_{\mathrm{homo}}$ updates $\mathcal R_\phi$ only through the perturbed prediction $\mathbf D_{\mathrm{pert}}$.

\paragraph{Overall Objective.}
The two terms form the geometric consistency objective. Combining them with SSHNet's intra-modal homography, modality-transfer, and feature-space objectives and dual-scale reconstruction gives
\begin{equation}
\begin{aligned}
\mathcal L_{\mathrm{consist}}
&=
\lambda_{\mathrm{trans}}\mathcal L_{\mathrm{trans}}
+
\lambda_{\mathrm{homo}}\mathcal L_{\mathrm{homo}},\\
\mathcal L_{\mathrm{total}}
&=
\mathcal L_{\mathrm{H,A}}
+
\mathcal L_{\mathrm{H,B}}
+
\mathcal L_{\mathrm T}
+
\mathcal L_{\mathrm{HF}}
\\
&\quad+
\lambda_{\mathrm{recon}}\mathcal L_{\mathrm{recon}}
+
\mathcal L_{\mathrm{consist}}.
\end{aligned}
\label{eq:overall-objective}
\end{equation}
The first four terms are inherited from SSHNet~\cite{yu2025sshnet}: $\mathcal L_{\mathrm{H,A}}$ and $\mathcal L_{\mathrm{H,B}}$ supervise homography estimation on synthetically warped intra-modal pairs from modalities A and B, respectively; $\mathcal L_{\mathrm T}$ supervises modality transfer using the homography-aligned target; and $\mathcal L_{\mathrm{HF}}$ enforces homography feature-space consistency. We retain SSHNet's split optimization and stop-gradient paths without modification. The coefficients $\lambda_{\mathrm{recon}}$, $\lambda_{\mathrm{trans}}$, and $\lambda_{\mathrm{homo}}$ balance the three additional objectives. The added $\mathcal L_{\mathrm{trans}}$ and $\mathcal L_{\mathrm{homo}}$ update the modality-transfer network $\mathcal T_\theta$ and RVHomo $\mathcal R_\phi$, respectively. Both consistency paths are used only during training and introduce no inference-time computation.

\subsection{Two-Stage Progressive Distillation}

\begin{figure}[t]
    \centering
    \includegraphics[width=\linewidth]{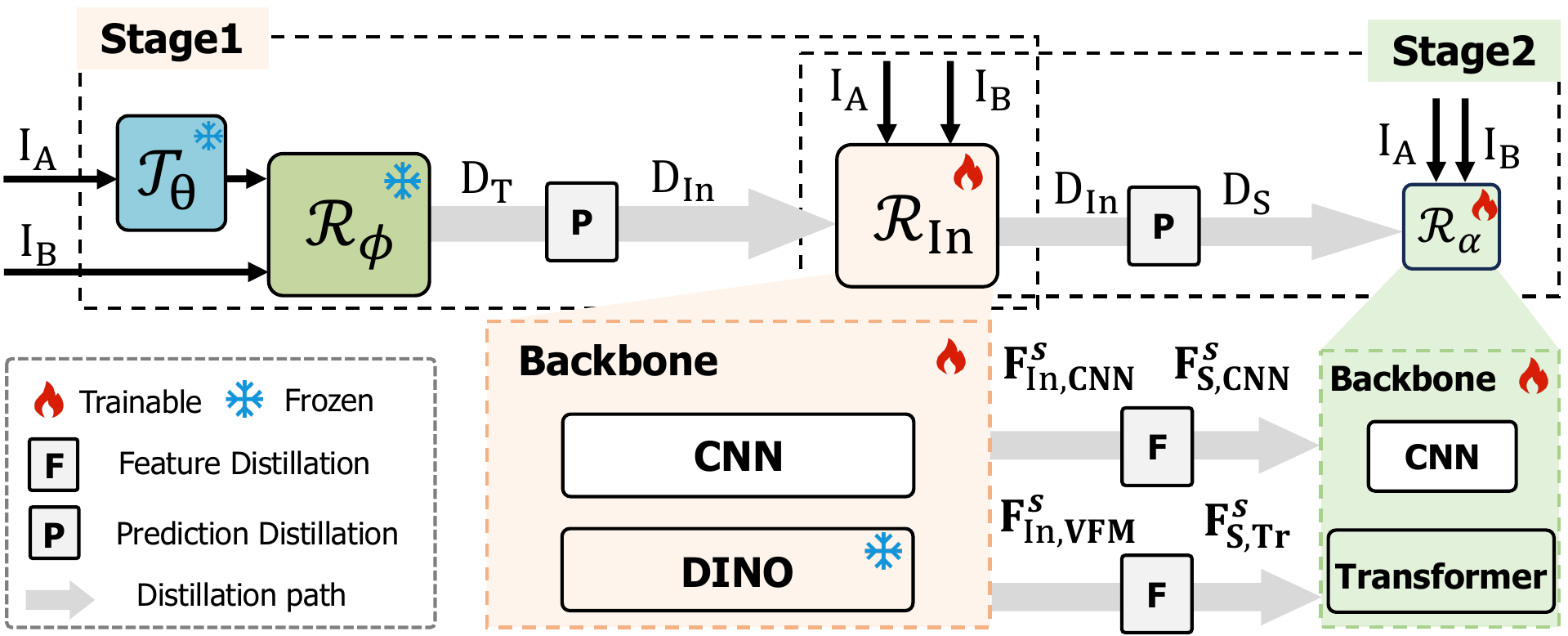}
    \caption{Two-stage progressive distillation. The complete GeoCo teacher transfers prediction supervision to the intermediate model $\mathcal R_{\mathrm{In}}$, which provides prediction and branch-wise feature supervision to the lightweight student $\mathcal R_\alpha$.}
    \label{fig:distillation}
\end{figure}

Knowledge distillation transfers predictions and representations to compact students~\cite{hinton2015distilling,romero2015fitnets,zagoruyko2017attention}. In \cref{fig:distillation}, the complete GeoCo teacher, comprising modality transfer $\mathcal T_\theta$ and RVHomo $\mathcal R_\phi$, supervises an intermediate model $\mathcal R_{\mathrm{In}}$ using SSHNet-D's prediction-distillation strategy~\cite{yu2025sshnet}. The intermediate model receives the original cross-modal pair, removing modality transfer. It then supplies prediction and feature supervision to RVHomo-D, the student $\mathcal R_\alpha$ whose lightweight Transformer replaces the VFM.

For prediction distillation, let $\mathbf D_{\mathrm{In}}\in\mathbb R^{4\times2}$ be the final four-corner prediction of $\mathcal R_{\mathrm{In}}$ and let $\{\mathbf D_{\mathrm S}^{(k)}\}_{k=1}^{K}$ be the $K$ recurrent predictions of $\mathcal R_\alpha$. Feature distillation is applied before branch fusion by aligning the student CNN branch with the intermediate CNN branch and the student Transformer branch with the intermediate VFM branch.

Let $\mathcal S=\{\tfrac14,\tfrac12\}$, $d(\cdot,\cdot)$ denote normalized feature discrepancy, and $\omega_s$ weight scale $s$. Prediction and modality-specific feature losses are combined as
\begin{equation}
\begin{aligned}
\mathcal L_{\mathrm{pred}}
&=
\sum_{k=1}^{K}\gamma^{K-k}
\operatorname{mean}\!\left(
\left\lvert
\mathbf D_{\mathrm S}^{(k)}
-
\sg(\mathbf D_{\mathrm{In}})
\right\rvert
\right),\\
\mathcal L_{\mathrm{feat},\mathrm A}
&=
\sum_{s\in\mathcal S}
\omega_s
\Big[
d\!\left(
\mathbf F_{\mathrm S,\mathrm{CNN},\mathrm A}^{s},
\sg(\mathbf F_{\mathrm{In},\mathrm{CNN},\mathrm A}^{s})
\right)
\\[-0.2em]
&\qquad\qquad+
d\!\left(
\mathbf F_{\mathrm S,\mathrm{Tr},\mathrm A}^{s},
\sg(\mathbf F_{\mathrm{In},\mathrm{VFM},\mathrm A}^{s})
\right)
\Big],\\
\mathcal L_{\mathrm{distill}}
&=
\lambda_{\mathrm{pred}}\mathcal L_{\mathrm{pred}}
+
\lambda_{\mathrm{feat}}
\left(
\mathcal L_{\mathrm{feat},\mathrm A}
+
\mathcal L_{\mathrm{feat},\mathrm B}
\right).
\end{aligned}
\label{eq:distillation-objective}
\end{equation}

Here, $\gamma$ is the recurrent prediction decay factor, and $\mathcal L_{\mathrm{feat},\mathrm B}$ is defined identically for modality $\mathrm B$. Lightweight projections align feature dimensions. At inference, only RVHomo-D is retained, without modality transfer, VFM extraction, or distillation-specific components.

%% file: sec/4_experiments.tex
\begin{table*}[t]
    \centering
    \small
    \caption{Cross-modal homography estimation (MACE). GM, OS, and VI denote GoogleMap, OPT-SAR, and VIS-IR. NC denotes non-convergence under the common training protocol. SSHNet variants exclude our consistency terms.}
    \label{tab:main-results}
    \begin{tabular*}{\textwidth}{@{\extracolsep{\fill}}lllrrrr@{}}
        \toprule
        Method family & Estimator / method & Publication & GM & OS & FSD & VI \\
        \midrule
        Generic unsupervised & UDHN~\cite{nguyen2018unsupervised} & RA-L 2018 & 22.84 & NC & NC & NC \\
        & CA-UDHN~\cite{zhang2020content} & ECCV 2020 & 24.61 & 24.76 & 24.86 & 24.55 \\
        & biHomE~\cite{koguciuk2021perceptual} & CVPRW 2021 & NC & NC & NC & NC \\
        & BasesHomo~\cite{ye2021motion} & ICCV 2021 & 24.49 & 26.72 & 25.08 & NC \\
        & SCPNet~\cite{zhang2024scpnet} & ECCV 2024 & 4.35 & NC & NC & NC \\
        \midrule
        AltO~\cite{song2024alto} & DHN~\cite{detone2016deep} & arXiv 2016 & 6.19 & 24.83 & NC & 12.24 \\
        & RAFT~\cite{teed2020raft} & ECCV 2020 & 3.10 & 11.38 & NC & 7.84 \\
        & IHN~\cite{cao2022iterative} & CVPR 2022 & 3.06 & 11.69 & NC & 6.83 \\
        & RHWF~\cite{cao2023recurrent} & CVPR 2023 & 3.49 & 13.97 & NC & 6.17 \\
        \midrule
        SSHNet~\cite{yu2025sshnet} & DHN~\cite{detone2016deep} & arXiv 2016 & 9.28 & 17.63 & 15.02 & 10.86 \\
        & MHN~\cite{le2020deep} & CVPR 2020 & 2.90 & 6.77 & 10.37 & 7.51 \\
        & IHN~\cite{cao2022iterative} & CVPR 2022 & 1.23 & 2.94 & 5.32 & 5.51 \\
        & RHWF~\cite{cao2023recurrent} & CVPR 2023 & 1.29 & 3.08 & 5.64 & 5.60 \\
        & MCNet~\cite{zhu2024mcnet} & CVPR 2024 & 1.22 & 2.87 & 6.06 & 6.22 \\
        \midrule
        GeoCo (ours) & \textbf{RVHomo} & -- & \textbf{0.65} & \textbf{1.95} & \textbf{4.32} & \textbf{4.55} \\
        GeoCo-D & \textbf{RVHomo-D} & -- & 0.71 & 2.04 & 4.77 & 4.72 \\
        \bottomrule
    \end{tabular*}
\end{table*}

\section{Experiments}
\label{sec:experiments}

\subsection{Experimental Settings}

\paragraph{Datasets and metric.}
We evaluate GeoCo on four cross-modal homography estimation benchmarks: GoogleMap~\cite{zhao2021deep}, OPT-SAR~\cite{li2022mcanet}, FSD~\cite{wen2025foundationstereo}, and VIS-IR~\cite{sun2022drone}. Following the standard corner-perturbation protocol for homography estimation~\cite{cao2022iterative,zhu2024mcnet,yu2025sshnet}, we generate cross-modal input pairs by applying synthetic homography perturbations to aligned cross-modal image pairs. Cross-modal transformation labels are used only for evaluation. We report mean average corner error (MACE) in pixels, with lower values indicating better accuracy.

\paragraph{Training and baselines.}
We compare our framework against generic unsupervised homography estimation methods and variants of AltO~\cite{song2024alto} and SSHNet~\cite{yu2025sshnet} equipped with different homography estimators. All methods are trained separately on each target dataset using shared data partitions and evaluation protocols. The SSHNet variants in Table~\ref{tab:main-results} exclude our consistency terms.

Further details on datasets, evaluation, training, and the NC criterion are provided in Supplementary Section~A, with consistency implementation described in Supplementary Section~C.

\subsection{Comparisons with Existing Methods}

\Cref{tab:main-results} reports the results on the four datasets. GeoCo achieves MACE values of 0.65, 1.95, 4.32, and 4.55 on GoogleMap, OPT-SAR, FSD, and VIS-IR, respectively. Compared with the best-performing baseline on each dataset---SSHNet-MCNet on GoogleMap and OPT-SAR, and SSHNet-IHN on FSD and VIS-IR---GeoCo reduces MACE by 46.7\%, 32.1\%, 18.8\%, and 17.4\%, respectively. Our distilled model, GeoCo-D, also outperforms all compared baselines across the four datasets, achieving MACE values of 0.71, 2.04, 4.77, and 4.72, respectively.

\begin{figure*}[t]
    \centering
    \includegraphics[trim=14.76bp 0 0 0,clip,width=\textwidth]{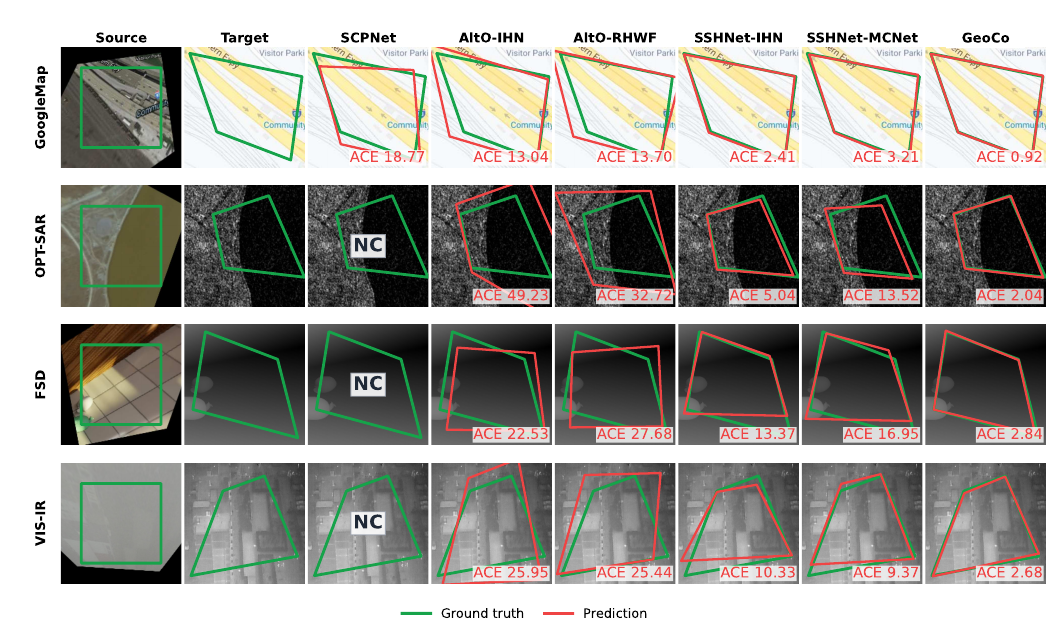}
    \caption{Qualitative comparison on four modality pairs. NC denotes non-convergence under the common training protocol.}
    \label{fig:qualitative}
\end{figure*}

\Cref{fig:qualitative} presents qualitative comparisons on the four datasets. The AltO variants show large corner errors, particularly in the OPT-SAR, FSD, and VIS-IR examples. SSHNet reduces these errors, while GeoCo further improves alignment with the ground truth. Our method achieves the lowest ACE in all four examples.

\subsection{Ablation Studies}

We conduct ablation studies to evaluate the contributions of geometric consistency constraints and VFM feature adaptation to GeoCo. Consistency ablations cover all four datasets, while comparisons of VFM feature-use strategies and architectural ablations of RVHomo are performed on OPT-SAR.

\begin{table}[!htbp]
    \centering
    \small
    \caption{Consistency ablation on four datasets.}
    \label{tab:dual-consistency}
    \setlength{\tabcolsep}{4pt}
    \begin{tabular}{@{}ccrrrr@{}}
        \toprule
        $\Ltrans$ & $\Lhomo$ & GM & OS & FSD & VI \\
        \midrule
        $\times$ & $\times$ & 0.93 & 2.38 & 4.52 & 4.75 \\
        $\times$ & $\checkmark$ & 0.82 & 2.19 & 4.35 & 4.70 \\
        $\checkmark$ & $\times$ & 0.69 & 2.04 & 4.44 & 4.63 \\
        $\checkmark$ & $\checkmark$ & \textbf{0.65} & \textbf{1.95} & \textbf{4.32} & \textbf{4.55} \\
        \bottomrule
    \end{tabular}
\end{table}

\begin{figure*}[t]
    \centering
    \begin{minipage}[t]{0.48\textwidth}
        \centering
        \includegraphics[width=0.32\linewidth]{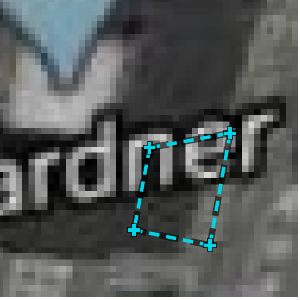}\hfill
        \includegraphics[width=0.32\linewidth]{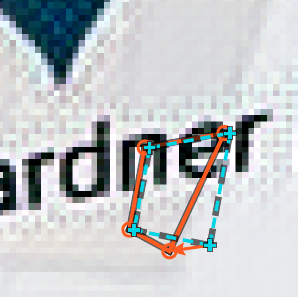}\hfill
        \includegraphics[width=0.32\linewidth]{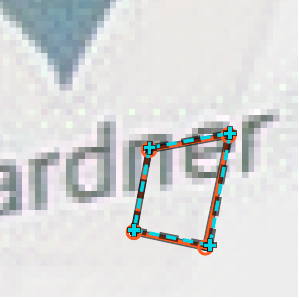}\\
        {\small\makebox[0.32\linewidth]{Source}\hfill\makebox[0.32\linewidth]{w/o $\Ltrans$}\hfill\makebox[0.32\linewidth]{w/ $\Ltrans$}}\\[2pt]
        {\small (a) Spatial preservation during modality transfer}
    \end{minipage}\hfill
    \begin{minipage}[t]{0.48\textwidth}
        \centering
        \includegraphics[width=0.32\linewidth]{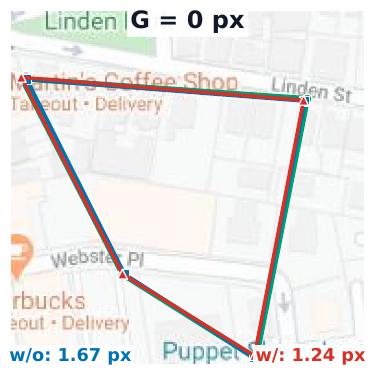}\hfill
        \includegraphics[width=0.32\linewidth]{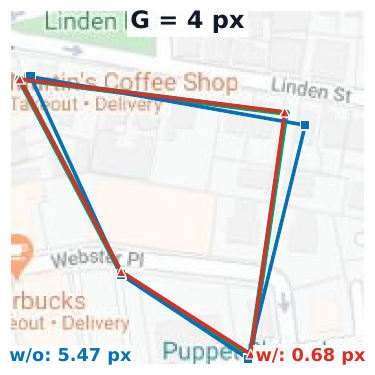}\hfill
        \includegraphics[width=0.32\linewidth]{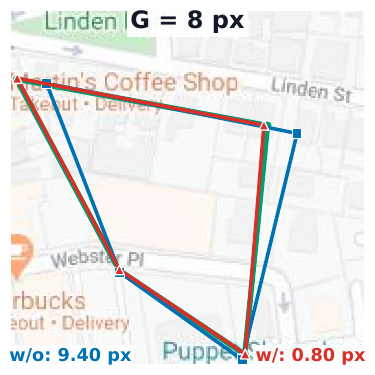}\\
        {\small\makebox[0.32\linewidth]{$G=0$ px}\hfill\makebox[0.32\linewidth]{$G=4$ px}\hfill\makebox[0.32\linewidth]{$G=8$ px}}\\[2pt]
        {\small (b) Prediction under source-view perturbations}
    \end{minipage}
    \caption{Effects of the two consistency constraints on GoogleMap examples. (a) Enlarged local regions: cyan dashed lines indicate reference topology; orange solid lines indicate translated topology. (b) Green circles denote ground truth, blue squares predictions without $\Lhomo$, and red triangles predictions with $\Lhomo$.}
    \label{fig:consistency-behavior}
\end{figure*}

\paragraph{Geometric consistency constraints.}
\Cref{tab:dual-consistency} shows that each consistency constraint reduces MACE on all four datasets, and combining them achieves the lowest error. Additional experiments in Supplementary Section~B show that the consistency losses improve accuracy across different homography estimators in most tested settings. \Cref{fig:consistency-behavior}(a) illustrates that modality-transfer consistency reduces geometric distortion during translation, better preserving the spatial structure of the source image. \Cref{fig:consistency-behavior}(b) shows that homography-prediction consistency keeps predictions closer to the ground truth under increasing source-view perturbations, indicating improved robustness to geometric perturbations on GoogleMap. Further analyses are provided in Supplementary Section~C.

\begin{table}[!htbp]
    \centering
    \small
    \caption{VFM feature-use strategies on OPT-SAR. In separated pathways, ``/'' distinguishes coarse and fine features.}
    \label{tab:vfm-feature-use}
    \renewcommand{\arraystretch}{1.12}
    \setlength{\tabcolsep}{3pt}
    \begin{tabular*}{\columnwidth}{@{\extracolsep{\fill}}llr@{}}
        \toprule
        Feature strategy & Feature design & MACE \\
        \midrule
        \multicolumn{3}{@{}l}{\underline{\emph{Supervised reference}}} \\
        GFNet & Cross-view + FPN & 2.498 \\
        \midrule
        \multicolumn{3}{@{}l}{\underline{\emph{Unsupervised variants}}} \\
        RoMa-style & DINO / VGG & 23.47 \\
        GFNet-style & Cross-view + FPN & 26.91 \\
        Separated DINO/CNN & DINO / local CNN & 24.37 \\
        \textbf{RVHomo} & GFU + top-down fusion & \textbf{2.38} \\
        \bottomrule
    \end{tabular*}
\end{table}

\paragraph{VFM feature-use strategies.}
\Cref{tab:vfm-feature-use} compares different strategies for incorporating VFM features into our unsupervised framework, with supervised \textbf{GFNet} included as a reference. Trained with synthetic cross-modal homography labels at the same input resolution, GFNet achieves a MACE of 2.498. For the unsupervised comparisons, we adapt the feature extraction strategies of RoMa and GFNet to our recurrent estimator. All unsupervised variants share the same training objectives, with both consistency losses disabled. \textbf{RoMa-style features} use deepest-layer DINOv2 features for coarse estimation and VGG features for refinement. \textbf{GFNet-style features} integrate deepest-layer DINOv2 features into a CNN feature pyramid after cross-view attention. \textbf{Separated DINO/CNN} uses deepest-layer DINOv2 features for coarse estimation and RVHomo's local CNN features for refinement, without fusion. \textbf{RVHomo} instead spatially adapts and aggregates hierarchical VFM features, then fuses them with local CNN features at both estimation scales. RVHomo achieves a MACE of 2.38, compared with 23.47--26.91 for the unsupervised alternatives. These results highlight the importance of designing VFM feature adaptation and fusion specifically for the unsupervised homography estimation framework. Implementation details and the supervised GFNet protocol are provided in Supplementary Sections~A and~B.

\begin{table}[!htbp]
    \centering
    \small
    \caption{RVHomo architecture ablation within GeoCo on OPT-SAR. NC denotes non-convergence under the common training protocol.}
    \label{tab:rvhomo-architecture}
    \begin{tabular}{@{}lr@{}}
        \toprule
        Configuration & MACE \\
        \midrule
        w/o VFM & 2.54 \\
        w/o GFU & 3.74 \\
        w/o top-down & NC \\
        \textbf{RVHomo} & \textbf{1.95} \\
        \bottomrule
    \end{tabular}
\end{table}

\paragraph{Architecture.}
With correlation construction and the recurrent estimator fixed, \cref{tab:rvhomo-architecture} varies VFM adaptation. The \textbf{w/o VFM} model forms both matching scales from the CNN pyramid. The \textbf{w/o GFU} model retains the four-level hierarchy and top-down pathway but uses bilinear resizing and convolutional projection. The \textbf{w/o top-down} model retains GFU-adapted levels and replaces sequential aggregation with parallel resizing and concatenation.

Removing the VFM branch raises MACE from 1.95 to 2.54; removing GFU raises it to 3.74. Without top-down aggregation, training does not converge under the common protocol. These results support learnable scale adaptation and hierarchical aggregation within the evaluated design, rather than direct insertion of coarse VFM features.

We further discuss the choice of DINOv2 feature layers and alternative adapter designs in Supplementary Section~D.

\subsection{Two-Stage Progressive Distillation}

\begin{table}[!htbp]
    \centering
    \small
    \caption{Distillation on OPT-SAR. All rows report the MACE of the final GeoCo-D model.}
    \label{tab:distill-targets}
    \begin{tabular}{@{}cccr@{}}
        \toprule
        Teacher & $\mathcal L_{\mathrm{pred}}$ & $\mathcal L_{\mathrm{feat}}$ & MACE \\
        \midrule
        GeoCo & $\checkmark$ & $\times$ & 2.80 \\
        GeoCo & $\checkmark$ & $\checkmark$ & 5.39 \\
        Intermediate & $\checkmark$ & $\times$ & 2.31 \\
        Intermediate & $\checkmark$ & $\checkmark$ & \textbf{2.04} \\
        \bottomrule
    \end{tabular}
\end{table}

\paragraph{Distillation ablation.}
\Cref{tab:distill-targets} compares distillation strategies for the final student on OPT-SAR. Direct prediction distillation yields 2.80 MACE, increasing to 5.39 with feature supervision. The teacher performs approximately intra-modal matching after modality transfer, whereas the student directly performs cross-modal matching. This difference suggests that forcing the student to reproduce the teacher's features may hinder the learning of representations suited to cross-modal matching. The intermediate model first learns to perform cross-modal matching through prediction distillation, then provides feature supervision for the student under the same matching task. Using the intermediate teacher yields 2.31 MACE with prediction supervision, which further improves to 2.04 when feature supervision is added. These results support the effectiveness of the two-stage progressive distillation strategy.

\begin{table}[!htbp]
    \centering
    \small
    \caption{Model complexity and measured efficiency. Parameters and FLOPs are reported in millions and billions, respectively; latency is in ms and memory in MiB.}
    \label{tab:distill-complexity}
    \setlength{\tabcolsep}{3pt}
    \begin{tabular}{@{}lrrrr@{}}
        \toprule
        Model & Params. & FLOPs & Latency & Memory \\
        \midrule
        GeoCo & 320.547 & 120.252 & 59.21 & 1387.60 \\
        Intermediate & 312.973 & 116.671 & 49.19 & 1347.53 \\
        \textbf{GeoCo-D} & \textbf{4.877} & \textbf{35.552} & \textbf{38.87} & \textbf{173.09} \\
        \bottomrule
    \end{tabular}
\end{table}

\paragraph{Model complexity.}
\Cref{tab:distill-complexity} shows that GeoCo-D reduces total parameters by 98.5\% and GFLOPs by 70.4\% relative to the complete GeoCo teacher. Architecture measurements on an RTX 4090 (FP32, batch size 1) give 34.3\% lower latency and 87.5\% lower peak allocated memory. Supplementary Section~E reports the complete protocol, trainable parameters, and timing variability.

%% file: sec/5_conclusion.tex
\section{Conclusion}
\label{sec:conclusion}

We presented GeoCo, a framework that makes better use of geometric information for unsupervised cross-modal homography estimation. Its RVHomo estimator adapts VFM features, while two geometric consistency constraints regularize modality transfer and homography prediction. Two-stage progressive distillation further produces GeoCo-D without modality transfer or online VFM extraction. Experiments on four cross-modal benchmarks demonstrate the effectiveness of these designs, our framework outperforming the compared baselines. These results highlight the value of jointly improving geometric representations and enforcing geometric consistency in unsupervised homography estimation. 
Future work will explore a unified model that generalizes across datasets, sensors, and modality combinations to support cross-modal alignment in more diverse imaging scenarios.

%% file: references.bib
@article{detone2016deep,
  author  = {DeTone, Daniel and Malisiewicz, Tomasz and Rabinovich, Andrew},
  title   = {Deep Image Homography Estimation},
  journal = {arXiv preprint arXiv:1606.03798},
  year    = {2016}
}

@inproceedings{teed2020raft,
  author    = {Teed, Zachary and Deng, Jia},
  title     = {{RAFT}: Recurrent All-Pairs Field Transforms for Optical Flow},
  booktitle = {Proceedings of the European Conference on Computer Vision},
  pages     = {402--419},
  year      = {2020}
}

@inproceedings{le2020deep,
  author    = {Le, Hoang and Liu, Feng and Zhang, Shu and Agarwala, Aseem},
  title     = {Deep Homography Estimation for Dynamic Scenes},
  booktitle = {Proceedings of the IEEE/CVF Conference on Computer Vision and Pattern Recognition},
  pages     = {7652--7661},
  year      = {2020}
}

@inproceedings{cao2022iterative,
  author    = {Cao, Si-Yuan and Hu, Jianxin and Sheng, Zehua and Shen, Hui-Liang},
  title     = {Iterative Deep Homography Estimation},
  booktitle = {Proceedings of the IEEE/CVF Conference on Computer Vision and Pattern Recognition},
  pages     = {1879--1888},
  year      = {2022}
}

@inproceedings{cao2023recurrent,
  author    = {Cao, Si-Yuan and Zhang, Runmin and Luo, Lun and Yu, Beinan and Sheng, Zehua and Li, Junwei and Shen, Hui-Liang},
  title     = {Recurrent Homography Estimation Using Homography-Guided Image Warping and Focus Transformer},
  booktitle = {Proceedings of the IEEE/CVF Conference on Computer Vision and Pattern Recognition},
  pages     = {9833--9842},
  year      = {2023}
}

@inproceedings{zhu2024mcnet,
  author    = {Zhu, Haokai and Cao, Si-Yuan and Hu, Jianxin and Zuo, Sitong and Yu, Beinan and Ying, Jiacheng and Li, Junwei and Shen, Hui-Liang},
  title     = {{MCNet}: Rethinking the Core Ingredients for Accurate and Efficient Homography Estimation},
  booktitle = {Proceedings of the IEEE/CVF Conference on Computer Vision and Pattern Recognition},
  pages     = {25932--25941},
  year      = {2024}
}

@article{nguyen2018unsupervised,
  author  = {Nguyen, Ty and Chen, Steven W. and Shivakumar, Shreyas S. and Taylor, Camillo J. and Kumar, Vijay},
  title   = {Unsupervised Deep Homography: A Fast and Robust Homography Estimation Model},
  journal = {IEEE Robotics and Automation Letters},
  volume  = {3},
  number  = {3},
  pages   = {2346--2353},
  year    = {2018},
  doi     = {10.1109/LRA.2018.2809549}
}

@inproceedings{zhang2020content,
  author    = {Zhang, Jirong and Wang, Chuan and Liu, Shuaicheng and Jia, Lanpeng and Ye, Nianjin and Wang, Jue and Zhou, Ji and Sun, Jian},
  title     = {Content-Aware Unsupervised Deep Homography Estimation},
  booktitle = {Proceedings of the European Conference on Computer Vision},
  pages     = {653--669},
  year      = {2020}
}

@inproceedings{koguciuk2021perceptual,
  author    = {Koguciuk, Daniel and Arani, Elahe and Zonooz, Bahram},
  title     = {Perceptual Loss for Robust Unsupervised Homography Estimation},
  booktitle = {Proceedings of the IEEE/CVF Conference on Computer Vision and Pattern Recognition Workshops},
  pages     = {4274--4283},
  year      = {2021}
}

@inproceedings{ye2021motion,
  author    = {Ye, Nianjin and Wang, Chuan and Fan, Haoqiang and Liu, Shuaicheng},
  title     = {Motion Basis Learning for Unsupervised Deep Homography Estimation with Subspace Projection},
  booktitle = {Proceedings of the IEEE/CVF International Conference on Computer Vision},
  pages     = {13117--13125},
  year      = {2021}
}

@inproceedings{zhang2024scpnet,
  author    = {Zhang, Runmin and Ma, Jun and Cao, Si-Yuan and Luo, Lun and Yu, Beinan and Chen, Shu-Jie and Li, Junwei and Shen, Hui-Liang},
  title     = {{SCPNet}: Unsupervised Cross-Modal Homography Estimation via Intra-Modal Self-Supervised Learning},
  booktitle = {Proceedings of the European Conference on Computer Vision},
  pages     = {460--477},
  year      = {2024}
}

@inproceedings{arar2020nemar,
  author    = {Arar, Moab and Ginger, Yiftach and Danon, Dov and Bermano, Amit H. and Cohen-Or, Daniel},
  title     = {Unsupervised Multi-Modal Image Registration via Geometry Preserving Image-to-Image Translation},
  booktitle = {Proceedings of the IEEE/CVF Conference on Computer Vision and Pattern Recognition},
  pages     = {13410--13419},
  year      = {2020}
}

@inproceedings{song2024alto,
  author    = {Song, Sanghyeob and Lew, Jaihyun and Jang, Hyemi and Yoon, Sungroh},
  title     = {Unsupervised Homography Estimation on Multimodal Image Pair via Alternating Optimization},
  booktitle = {Advances in Neural Information Processing Systems},
  volume    = {37},
  pages     = {61306--61327},
  year      = {2024}
}

@inproceedings{yu2025sshnet,
  author    = {Yu, Junchen and Cao, Si-Yuan and Zhang, Runmin and Zhang, Chenghao and Yu, Zhu and Chen, Shujie and Yang, Bailin and Shen, Hui-Liang},
  title     = {{SSHNet}: Unsupervised Cross-Modal Homography Estimation via Problem Reformulation and Split Optimization},
  booktitle = {Proceedings of the IEEE/CVF Conference on Computer Vision and Pattern Recognition},
  pages     = {16685--16694},
  year      = {2025}
}

@inproceedings{you2026generalized,
  author    = {You, Jinkun and Cheng, Jiaxin and Zhang, Jie and Zhou, Yicong},
  title     = {Towards Generalized Multimodal Homography Estimation},
  booktitle = {Proceedings of the IEEE/CVF Conference on Computer Vision and Pattern Recognition},
  pages     = {8408--8417},
  year      = {2026}
}

@inproceedings{zhang2025gfnet,
  author    = {Zhang, Kaining and Deng, Yuxin and Ma, Jiayi and Favaro, Paolo},
  title     = {Adapting Dense Matching for Homography Estimation with Grid-based Acceleration},
  booktitle = {Proceedings of the IEEE/CVF Conference on Computer Vision and Pattern Recognition},
  pages     = {6294--6303},
  year      = {2025}
}

@article{yang2025distillmatch,
  author  = {Yang, Meng and Fan, Fan and Li, Zizhuo and Deng, Songchu and Ma, Yong and Ma, Jiayi},
  title   = {{DistillMatch}: Leveraging Knowledge Distillation from Vision Foundation Model for Multimodal Image Matching},
  journal = {arXiv preprint arXiv:2509.16017},
  year    = {2025}
}

@article{oquab2023dinov2,
  author  = {Oquab, Maxime and Darcet, Timoth{\'e}e and Moutakanni, Th{\'e}o and Vo, Huy V. and Szafraniec, Marc and Khalidov, Vasil and Fernandez, Pierre and Haziza, Daniel and Massa, Francisco and El-Nouby, Alaaeldin and Assran, Mahmoud and Ballas, Nicolas and Galuba, Wojciech and Howes, Russell and Huang, Po-Yao and Li, Shang-Wen and Misra, Ishan and Rabbat, Michael and Sharma, Vasu and Synnaeve, Gabriel and Xu, Hu and J{\'e}gou, Herv{\'e} and Mairal, Julien and Labatut, Patrick and Joulin, Armand and Bojanowski, Piotr},
  title   = {{DINOv2}: Learning Robust Visual Features without Supervision},
  journal = {Transactions on Machine Learning Research},
  year    = {2024}
}

@inproceedings{song2024dinoreg,
  author    = {Song, Xinrui and Xu, Xuanang and Yan, Pingkun},
  title     = {{DINO-Reg}: General Purpose Image Encoder for Training-Free Multi-Modal Deformable Medical Image Registration},
  booktitle = {Medical Image Computing and Computer Assisted Intervention},
  volume    = {15002},
  pages     = {608--617},
  publisher = {Springer Nature Switzerland},
  year      = {2024}
}

@inproceedings{shi2016realtime,
  author    = {Shi, Wenzhe and Caballero, Jose and Husz{\'a}r, Ferenc and Totz, Johannes and Aitken, Andrew P. and Bishop, Rob and Rueckert, Daniel and Wang, Zehan},
  title     = {Real-Time Single Image and Video Super-Resolution Using an Efficient Sub-Pixel Convolutional Neural Network},
  booktitle = {Proceedings of the IEEE Conference on Computer Vision and Pattern Recognition},
  pages     = {1874--1883},
  year      = {2016}
}

@article{hinton2015distilling,
  author  = {Hinton, Geoffrey and Vinyals, Oriol and Dean, Jeff},
  title   = {Distilling the Knowledge in a Neural Network},
  journal = {arXiv preprint arXiv:1503.02531},
  year    = {2015}
}

@inproceedings{romero2015fitnets,
  author    = {Romero, Adriana and Ballas, Nicolas and Kahou, Samira Ebrahimi and Chassang, Antoine and Gatta, Carlo and Bengio, Yoshua},
  title     = {{FitNets}: Hints for Thin Deep Nets},
  booktitle = {International Conference on Learning Representations},
  year      = {2015}
}

@inproceedings{johnson2016perceptual,
  author    = {Johnson, Justin and Alahi, Alexandre and Fei-Fei, Li},
  title     = {Perceptual Losses for Real-Time Style Transfer and Super-Resolution},
  booktitle = {Proceedings of the European Conference on Computer Vision},
  pages     = {694--711},
  year      = {2016}
}

@inproceedings{zhao2021deep,
  author    = {Zhao, Yiming and Huang, Xinming and Zhang, Ziming},
  title     = {Deep {Lucas-Kanade} Homography for Multimodal Image Alignment},
  booktitle = {Proceedings of the IEEE/CVF Conference on Computer Vision and Pattern Recognition},
  pages     = {15950--15959},
  year      = {2021}
}

@article{li2022mcanet,
  author  = {Li, Xue and Zhang, Guo and Cui, Hao and Hou, Shasha and Wang, Shunyao and Li, Xin and Chen, Yujia and Li, Zhijiang and Zhang, Li},
  title   = {{MCANet}: A Joint Semantic Segmentation Framework of Optical and {SAR} Images for Land Use Classification},
  journal = {International Journal of Applied Earth Observation and Geoinformation},
  volume  = {106},
  pages   = {102638},
  year    = {2022},
  doi     = {10.1016/j.jag.2021.102638}
}

@inproceedings{wen2025foundationstereo,
  author    = {Wen, Bowen and Trepte, Matthew and Aribido, Joseph and Kautz, Jan and Gallo, Orazio and Birchfield, Stan},
  title     = {{FoundationStereo}: Zero-Shot Stereo Matching},
  booktitle = {Proceedings of the IEEE/CVF Conference on Computer Vision and Pattern Recognition},
  pages     = {5249--5260},
  year      = {2025}
}

@article{sun2022drone,
  author  = {Sun, Yiming and Cao, Bing and Zhu, Pengfei and Hu, Qinghua},
  title   = {Drone-Based {RGB}-Infrared Cross-Modality Vehicle Detection via Uncertainty-Aware Learning},
  journal = {IEEE Transactions on Circuits and Systems for Video Technology},
  volume  = {32},
  number  = {10},
  pages   = {6700--6713},
  year    = {2022},
  doi     = {10.1109/TCSVT.2022.3168279}
}

@inproceedings{hong2022homogan,
  author    = {Hong, Mingbo and Lu, Yuhang and Ye, Nianjin and Lin, Chunyu and Zhao, Qijun and Liu, Shuaicheng},
  title     = {Unsupervised Homography Estimation with Coplanarity-Aware {GAN}},
  booktitle = {Proceedings of the IEEE/CVF Conference on Computer Vision and Pattern Recognition},
  pages     = {17663--17672},
  year      = {2022}
}

@inproceedings{liu2023geoformer,
  author    = {Liu, Jiazhen and Li, Xirong},
  title     = {Geometrized Transformer for Self-Supervised Homography Estimation},
  booktitle = {Proceedings of the IEEE/CVF International Conference on Computer Vision},
  pages     = {9556--9565},
  year      = {2023}
}

@article{heinrich2012mind,
  author  = {Heinrich, Mattias P. and Jenkinson, Mark and Bhushan, Manav and Matin, Tahreema and Gleeson, Fergus V. and Brady, Michael and Schnabel, Julia A.},
  title   = {{MIND}: Modality Independent Neighbourhood Descriptor for Multi-Modal Deformable Registration},
  journal = {Medical Image Analysis},
  volume  = {16},
  number  = {7},
  pages   = {1423--1435},
  year    = {2012},
  doi     = {10.1016/j.media.2012.05.008}
}

@inproceedings{kim2015dasc,
  author    = {Kim, Seungryong and Min, Dongbo and Ham, Bumsub and Ryu, Seungchul and Do, Minh N. and Sohn, Kwanghoon},
  title     = {{DASC}: Dense Adaptive Self-Correlation Descriptor for Multi-Modal and Multi-Spectral Correspondence},
  booktitle = {Proceedings of the IEEE Conference on Computer Vision and Pattern Recognition},
  pages     = {2103--2112},
  year      = {2015}
}

@article{li2020rift,
  author  = {Li, Jiayuan and Hu, Qingwu and Ai, Mingyao},
  title   = {{RIFT}: Multi-Modal Image Matching Based on Radiation-Variation Insensitive Feature Transform},
  journal = {IEEE Transactions on Image Processing},
  volume  = {29},
  pages   = {3296--3310},
  year    = {2020},
  doi     = {10.1109/TIP.2019.2959244}
}

@inproceedings{pielawski2020comir,
  author    = {Pielawski, Nicolas and Wetzer, Elisabeth and {\"O}fverstedt, Johan and Lu, Jiahao and W{\"a}hlby, Carolina and Lindblad, Joakim and Sladoje, Nata{\v{s}}a},
  title     = {{CoMIR}: Contrastive Multimodal Image Representation for Registration},
  booktitle = {Advances in Neural Information Processing Systems},
  volume    = {33},
  pages     = {18433--18444},
  year      = {2020}
}

@inproceedings{caron2021dino,
  author    = {Caron, Mathilde and Touvron, Hugo and Misra, Ishan and J{\'e}gou, Herv{\'e} and Mairal, Julien and Bojanowski, Piotr and Joulin, Armand},
  title     = {Emerging Properties in Self-Supervised Vision Transformers},
  booktitle = {Proceedings of the IEEE/CVF International Conference on Computer Vision},
  pages     = {9650--9660},
  year      = {2021}
}

@inproceedings{weinzaepfel2022croco,
  author    = {Weinzaepfel, Philippe and Leroy, Vincent and Lucas, Thomas and Br{\'e}gier, Romain and Cabon, Yohann and Arora, Vaibhav and Antsfeld, Leonid and Chidlovskii, Boris and Csurka, Gabriela and Revaud, J{\'e}r{\^o}me},
  title     = {{CroCo}: Self-Supervised Pre-Training for {3D} Vision Tasks by Cross-View Completion},
  booktitle = {Advances in Neural Information Processing Systems},
  volume    = {35},
  pages     = {3502--3516},
  year      = {2022}
}

@inproceedings{tang2023dift,
  author    = {Tang, Luming and Jia, Menglin and Wang, Qianqian and Phoo, Cheng Perng and Hariharan, Bharath},
  title     = {Emergent Correspondence from Image Diffusion},
  booktitle = {Advances in Neural Information Processing Systems},
  volume    = {36},
  pages     = {1363--1389},
  year      = {2023}
}

@inproceedings{zhang2023twofeatures,
  author    = {Zhang, Junyi and Herrmann, Charles and Hur, Junhwa and Cabrera, Luisa Polania and Jampani, Varun and Sun, Deqing and Yang, Ming-Hsuan},
  title     = {A Tale of Two Features: {Stable Diffusion} Complements {DINO} for Zero-Shot Semantic Correspondence},
  booktitle = {Advances in Neural Information Processing Systems},
  volume    = {36},
  pages     = {45533--45547},
  year      = {2023}
}

@inproceedings{truong2021warp,
  author    = {Truong, Prune and Danelljan, Martin and Yu, Fisher and Van Gool, Luc},
  title     = {Warp Consistency for Unsupervised Learning of Dense Correspondences},
  booktitle = {Proceedings of the IEEE/CVF International Conference on Computer Vision},
  pages     = {10346--10356},
  year      = {2021}
}

@inproceedings{zagoruyko2017attention,
  author    = {Zagoruyko, Sergey and Komodakis, Nikos},
  title     = {Paying More Attention to Attention: Improving the Performance of Convolutional Neural Networks via Attention Transfer},
  booktitle = {International Conference on Learning Representations},
  year      = {2017}
}

@inproceedings{edstedt2024roma,
  author    = {Edstedt, Johan and Sun, Qiyu and B{\"o}kman, Georg and Wadenb{\"a}ck, M{\r a}rten and Felsberg, Michael},
  title     = {{RoMa}: Robust Dense Feature Matching},
  booktitle = {Proceedings of the IEEE/CVF Conference on Computer Vision and Pattern Recognition},
  pages     = {19790--19800},
  year      = {2024}
}

@inproceedings{zhang2026dcflow,
  author    = {Zhang, Runmin and Wang, Jialiang and Cao, Si-Yuan and Yu, Zhu and Yu, Junchen and Zhang, Guangyi and Shen, Hui-Liang},
  title     = {Rethinking Unsupervised Cross-modal Flow Estimation: Learning from Decoupled Optimization and Consistency Constraint},
  booktitle = {International Conference on Learning Representations},
  year      = {2026}
}
